\documentclass[10pt,twocolumn,letterpaper]{article}

\usepackage[pagenumbers]{cvpr} 

\usepackage{graphicx}
\usepackage{amsmath}
\usepackage{amssymb}
\usepackage{amsthm}
\usepackage{booktabs}
\usepackage{svg}
\usepackage{xcolor}
\usepackage[ruled,linesnumbered]{algorithm2e}
\usepackage{mathtools}
\usepackage{nicefrac}
\usepackage{soul}
\usepackage{multirow}
\usepackage{subcaption}
\usepackage{threeparttable}
\usepackage{enumitem}
\usepackage{tikz}
\usepackage{graphicx}
\usepackage{svg}
\usetikzlibrary{calc}
\usepackage{caption}  
\usepackage{array}
\usepackage{blindtext}
\usepackage{xifthen} 

\let\svthefootnote\thefootnote

\newcommand\blankfootnote[1]{%
  \let\thefootnote\relax\footnotetext{#1}%
  \let\thefootnote\svthefootnote%
}
\let\svfootnote\footnote
\renewcommand\footnote[2][?]{%
  \if\relax#1\relax%
    \blankfootnote{#2}%
  \else%
    \if?#1\svfootnote{#2}\else\svfootnote[#1]{#2}\fi%
  \fi
}

\definecolor{generate-color}{RGB}{20,0,20}
\definecolor{DarkGray}{RGB}{85, 85, 85}
\definecolor{customyellow}{HTML}{FFFF00}
\definecolor{charcoal}{HTML}{36454F} 

\newsavebox{\mycircle}

\SetCommentSty{mycommfont}
\SetKwComment{Comment}{// }{ }
\SetKwInput{KwInput}{Input}  

\newtheorem{definition}{Definition}
\newtheorem{lemma}{Lemma}
\newtheorem{theorem}{Theorem}

\newtheorem{proposition}{Proposition}

\newcommand{\yellowcircle}[1]{%
    \tikz[baseline=(char.base)]{
        \node[shape=circle,fill=yellow,draw=black,inner sep=1pt] (char) {\scriptsize\textbf{#1}};%
    }%
}

\definecolor{darktealblue}{RGB}{0, 124, 114}
\definecolor{darkbrownorange}{RGB}{163, 51, 0}
\newcommand{\ch}[1]{%
    \ifthenelse{\isempty{#1}}{
        \smaller\makebox[2.7em][r]{}
    }{%
    \ifdim #1 pt > 0 pt
        \textcolor{darktealblue}{\smaller\makebox[2.7em][r]{$+#1$}}%
    \else
        \textcolor{darkbrownorange}{\smaller\makebox[2.7em][r]{$#1$}}%
    \fi
    }
}

\newcommand{\cml}[1]{\multicolumn{1}{l}{#1}}

\definecolor{cvprblue}{rgb}{0.21,0.49,0.74}
\usepackage[pagebackref,breaklinks,colorlinks,allcolors=cvprblue]{hyperref}

\def\paperID{13683} 
\def\confName{CVPR}
\def\confYear{2025}

\title{Activity Recognition on Avatar-Anonymized Datasets \\
with Masked Differential Privacy}

\author{David Schneider$^{1,\star}$ \and
Sina Sajadmanesh$^{2}$ \and
Vikash Sehwag$^{2}$ \and
Saquib Sarfraz$^{1}$ \and
Rainer Stiefelhagen$^{1}$ \and
Lingjuan Lyu$^{2}$ \and
Vivek Sharma$^{2,\dag}$}

\newcommand{\Index}{i}
\newcommand{\TrainingIter}{t}
\newcommand{\NumEpochs}{T}
\newcommand{\DatasetSize}{N}

\newcommand{\GradientClipThreshold}{C}
\newcommand{\DPNoiseVariance}{\sigma}
\newcommand{\LearningRate}{\eta}
\newcommand{\BatchSize}{B}

\newcommand{\RawVideoSample}{\mathbf{x}}
\newcommand{\SampleWithPersonsRemoved}{\dot{\mathbf{x}}}
\newcommand{\AnonymizedVideoSample}{\tilde{\mathbf{x}}}
\newcommand{\ProtectedDataset}{\Dataset_{\text{p}}}
\newcommand{\UnprotectedDataset}{\Dataset_{\text{u}}}
\newcommand{\ModelParams}{\mathbf{\Theta}}

\newcommand{\LabelVector}{\mathbf{y}}

\newcommand{\GradientPriv}[1]{g^{#1}_{\text{pr}}}
\newcommand{\GradientPrivClipped}[1]{\bar{g}^{#1}_{\text{pr}}}

\newcommand{\GradientUtil}[1]{g^{#1}_{\text{pu}}}
\newcommand{\GradientTotal}[1]{g^{#1}_{\text{total}}}
\newcommand{\RegionsPerson}{\mathbf{q}}
\newcommand{\RegionsAnonymized}{\mathbf{r}}
\newcommand{\Pose}{\mathbf{p}}
\newcommand{\RenderedPerson}{\mathbf{h}}

\newcommand{\RawVideoDataset}{\mathcal{D}}

\newcommand{\AnonymizedVideoDataset}{\mathcal{D}''}

\newcommand{\BackgroundTokens}{\mathcal{T}_\text{pr}}
\newcommand{\PersonTokens}{\mathcal{T}_\text{pu}}

\newcommand{\TaskModel}{\mathcal{M}_\Theta}

\newcommand{\PersonSegmentation}{f_\text{seg}}
\newcommand{\PoseEstimation}{f_\text{pos}}
\newcommand{\Rendering}{f_\text{ren}}
\newcommand{\VideoInpainting}{f_\text{inp}}

\newcommand{\Tokenizer}{f_\text{tok}}
\newcommand{\LossFunction}{\mathcal{L}}

\newcommand{\Batch}{\mathcal{B}}
\newcommand{\SamplingProb}{\nicefrac{\BatchSize}{\DatasetSize}}

\newcommand{\Record}{r}
\newcommand{\MaskFunction}{f_\text{pr}}

\newcommand{\SensitiveSet}{s}
\newcommand{\NumTokens}{K}
\newcommand{\Adjacent}{\stackrel{\text{m}}{\sim}}

\begin{document}
\maketitle

\newcommand\nnfootnote[1]{%
  \begin{NoHyper}
  \renewcommand\thefootnote{}\footnote{#1}%
  \addtocounter{footnote}{-1}%
  \end{NoHyper}
}

\nnfootnote{Preprint. $^{1}$Karlsruhe Institute of Technology $^{2}$Sony AI, Zürich $^{\star}$Work done while interning at Sony AI.  $^{\dag}$VS started and led the project. Correspondence to: viveksharma@sony.com.} 

\begin{abstract}
Privacy-preserving computer vision is an important emerging problem in machine learning and artificial intelligence.
Prevalent methods tackling this problem use differential privacy (DP) or obfuscation techniques to protect the privacy of individuals. In both cases, the utility of the trained model is sacrificed heavily in this process. In this work, we present an anonymization pipeline that replaces sensitive human subjects in video datasets with synthetic avatars within context, employing a combined rendering and stable diffusion-based strategy. 
Additionally we propose masked differential privacy ({MaskDP}) to protect non-anonymized but privacy sensitive background information. MaskDP allows for controlling sensitive regions where differential privacy is applied, in contrast to applying DP on the entire input.
This combined methodology provides strong privacy protection while minimizing the usual performance penalty of privacy preserving methods.
Experiments on multiple challenging action recognition datasets demonstrate that our proposed techniques result in better utility-privacy trade-offs compared to standard differentially private training in the especially demanding $\epsilon<1$ regime.
\end{abstract}
\vspace{-1em}

\section{Introduction}\label{sec:intro}

\begin{figure}[!tb]
  \centering
  \includegraphics[width=0.47\textwidth]{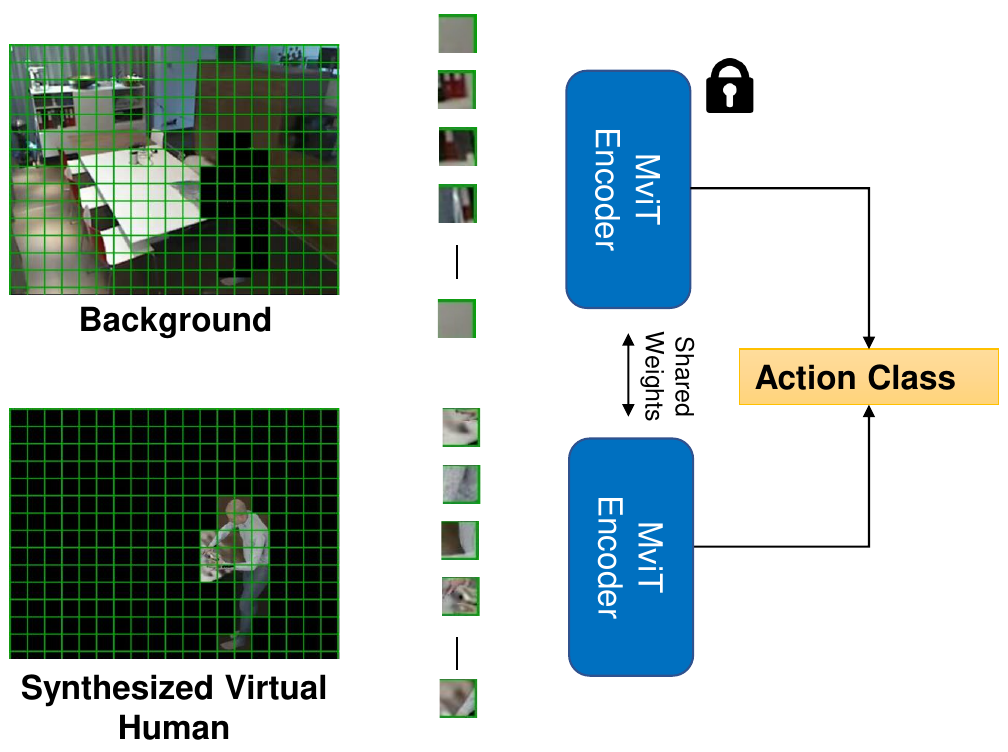}
  \caption{
In our experiments we apply MaskDP on anonymized datasets containing non-sensitive synthetic humans. Only real-world data is considered private and needs protection.}
  \label{fig:training_scheme}
\vspace{-1em}
\end{figure}

\newcommand{\numberedcircle}[4][center]{
    \filldraw[fill=yellow, draw=black, thin] (#2, #3) circle (0.17cm);
    \node[anchor=center] at (#2, #3) {\small\textbf{#4}};
}

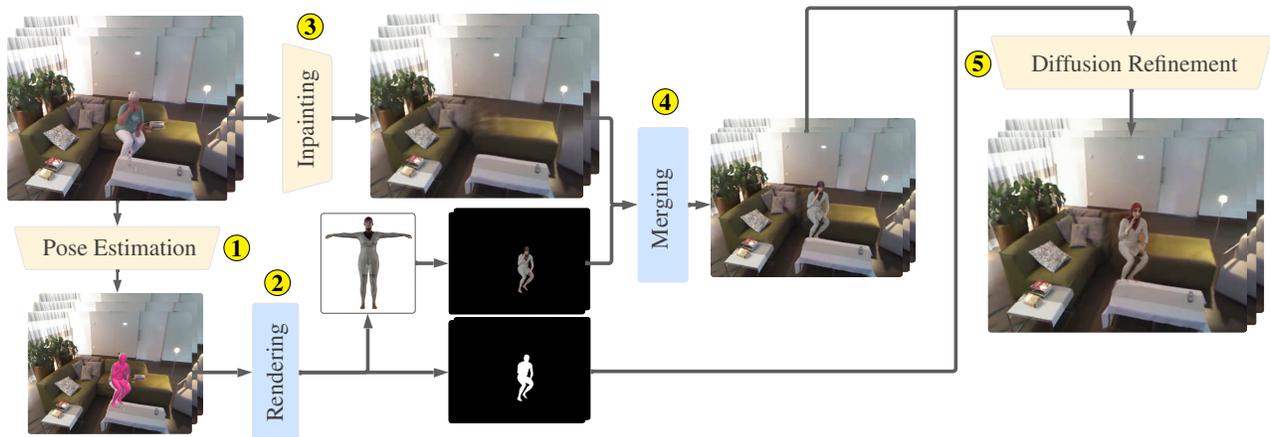
\begin{figure*}[t]
    \centering
    \begin{tikzpicture}
        \node[anchor=south west,inner sep=0pt] (image) at (0,0) {\includegraphics[width=\textwidth]{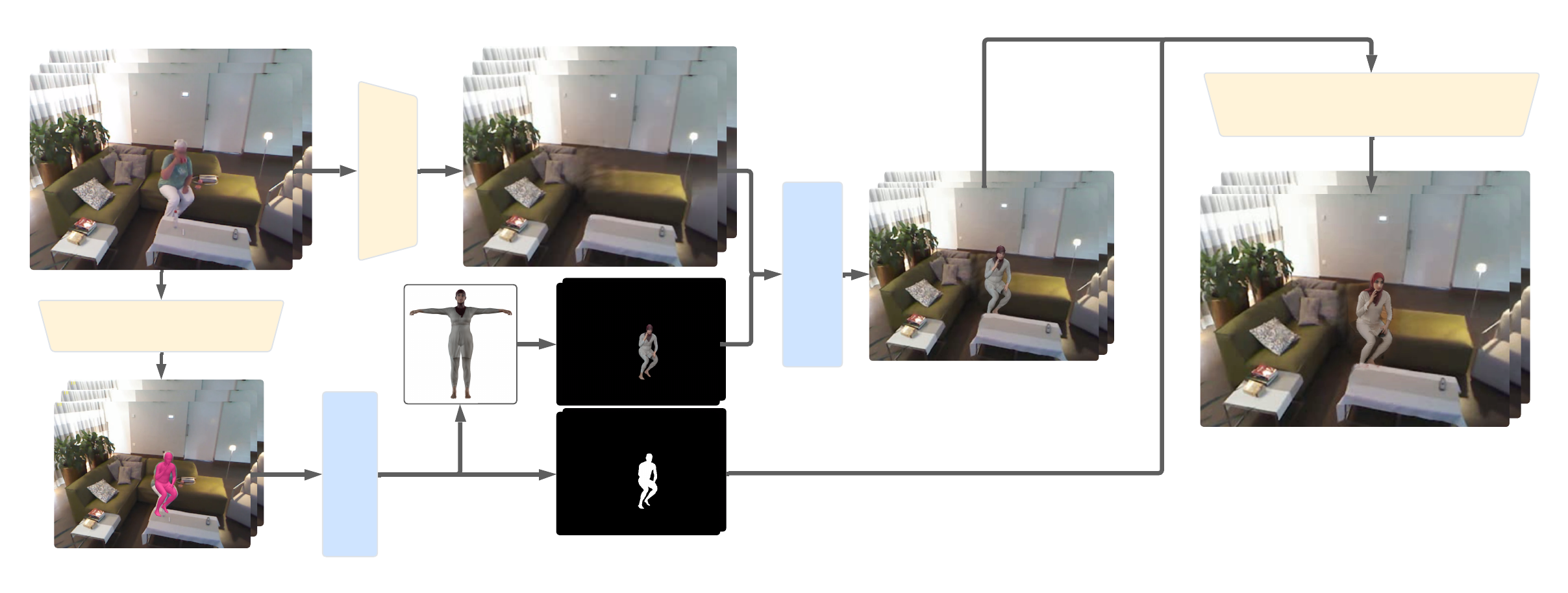}};

    \begin{scope}[x={(image.south east)}, y={(image.north west)}]
        \numberedcircle{0.194}{0.45}{1}
        \numberedcircle{0.225}{0.38}{2}
        \numberedcircle{0.25}{0.895}{3}
        \numberedcircle{0.52}{0.744}{4}
        \numberedcircle{0.758}{0.82}{5}

        \begin{scope}[anchor=center, text=gray!150]
            \node            at (.105, .451) {\small{Pose Estimation}};
            \node[rotate=90] at (.225, .2) {\small{Rendering}};
            \node[rotate=90] at (.2484, .71) {\small{Inpainting}};
            \node[rotate=90] at (.5177, .53) {\small{Merging}};
            \node            at (.877, .821) {\small{Diffusion Refinement}};
        \end{scope}
    \end{scope}
        
    \end{tikzpicture}
    \vspace{-0.6cm} 
\caption{Overview of our video anonymization framework. The process includes: \protect\yellowcircle{1} 3D pose estimation, \protect\yellowcircle{2} Synthetic human rendering using SMPL-H, \protect\yellowcircle{3} Background inpainting to remove original persons, \protect\yellowcircle{4} Merging of synthetic avatar and inpainted background, and \protect\yellowcircle{5} Diffusion-based refinement. This pipeline effectively anonymizes individuals while preserving video quality and context.}
    \label{fig:framework}
    \vspace{-0.2cm}
\end{figure*}

In the past two decades, human activity recognition~\cite{wang2013action,Karpathy,i3d,diba2020large} in videos has gained significant attention for its wide applications in video retrieval, behavior analysis, surveillance, and video understanding tasks. Despite notable progress in this field, concerns about data privacy have escalated as videos often contain sensitive information, such as human faces and other personal attributes. Safeguarding individuals' privacy rights in video understanding is not only a legal and ethical obligation but also a necessity to build up user trust under regulatory frameworks, such as GDPR~\cite{gdpr}, CCPA~\cite{ccpa}, and EU AI Act~\cite{euaiact}.
Without rigorous privacy measures, there are substantial risks of misuse, 
emphasizing the need for effective privacy-preserving techniques.

Recent efforts to integrate privacy-preserving measures, either with formal or empirical guarantees, into computer vision algorithms operating on sensitive datasets can be categorized into three types: (i) obfuscation-based approaches~\cite{dave2022spact,fioresi2023ted}, which hide semantic information while retaining utility information through minimax optimization; (ii) training on purely synthetic data which aims to mimic real-world patterns~\cite{de2017procedural,synadl,da2022dual,varol21_surreact}; and (iii) training with Differential Privacy~(DP)~\cite{luo2023differentially}, which involves adding noise to input data, output labels, or model parameters. 

While obfuscation-based methods often lack privacy guarantees, the usage of synthetic data or differential privacy results in significant utility penalties. These approaches treat entire images or videos as protected units, limiting the flexibility to apply privacy measures selectively to specific areas or elements within the data.

Our framework addresses these challenges through a two-stage process. The first stage involves a comprehensive anonymization pipeline that combines 3D avatar rendering with diffusion models to replace real human figures with synthetic ones while preserving the original video's background and context composition.
%

This approach offers several advantages: it provides pixel-wise information on synthetic data presence, preserves structural accuracy of movements, enhances realism, and maintains vital context which is difficult to generate synthetically. For many applications, such as activity recordings in public spaces, this method provides sufficient anonymity by completely replacing the original subject.

The second stage of our framework introduces Masked Differential Privacy (MaskDP), a novel relaxation of standard Differential Privacy. MaskDP leverages the pixel-wise knowledge about the presence of synthesized regions provided by our anonymization pipeline, allowing for selective application of additional privacy measures by decomposing data samples into private and public tokens. While public tokens carry no sensitive information, private tokens require protection.
MaskDP addresses this by exclusively applying differential privacy to private tokens, allowing our approach to achieve higher utility compared to standard DP, which would indiscriminately protect the entire video.
We demonstrate this capability by applying MaskDP on partially synthetic human action videos where regions containing synthetic data are considered non-sensitive, while remaining spatio-temporal regions are considered private.

This approach explicitly does not provide differential privacy guarantees for full video samples. Much rather, MaskDP only aims to provide differential privacy guarantees for explicitly defined privacy sensitive spatio-temporal regions, while either relying on other privacy techniques in non-protected areas, such as anonymization techniques, or declaring such areas to be non sensitive as would be the case for background information of recordings in public spaces.

Overall, we present the following key contributions:
\vspace{1em}

\begin{itemize}[topsep=4pt,itemsep=2pt,partopsep=2pt, parsep=2pt] 

    \item A framework for high quality fully automatic person anonymization at scale, minimizing the synthetic-to-real utility gap on major action recognition datasets, thereby demonstrating its usefulness on real world applications.
    \item Masked Differential Privacy (MaskDP), a novel relaxation of standard DP which leverages knowledge about private and public spatio-temporal areas as well as a theoretical analysis of its formal privacy guarantees, establishing connections with standard differential privacy.  
    \item Extensive experiments with differential privacy under various $\epsilon$ privacy budgets, outlining the DP utility-privacy trade-off for human activity recognition (HAR) in detail.
    \item An analysis of MaskDP on mixed synthetic-and-real datasets with anonymized humans, improving over standard DP in the strongly private low $\epsilon$ regime.
\end{itemize}

\section{Related Work}\label{sec:related}
\paragraph{Trasformer-based Action recognition.}
  
Recent advancements have highlighted the versatility of Transformers~\cite{vaswani2017attention}, initially prominent in natural language processing and subsequently applied to vision tasks such as image classification~\cite{dosovitskiy2020image}, video captioning~\cite{seo2022end}, and multimodal representation learning~\cite{luo2022clip4clip,lin2022eclipse}, extending to video classification~\cite{sun2019videobert,arnab2021vivit,yan2022multiview}. In action recognition, models like STAM~\cite{sharir2021image}, ViVit~\cite{arnab2021vivit}, and MVT~\cite{yan2022multiview} leverage video frames for analysis, while STCA~\cite{diba2023spatio} focuses on contextual dynamics for temporal insights. 

\paragraph{Privacy preserving action recognition.}
 
Recent privacy preservation techniques can be categorized into three major groups: downsampling-based approaches, obfuscation-based approaches, and adversarial training-based approaches. Downsampling based approaches anonymize data via low-resolution inputs, as shown by Chou et al.~\cite{chou2018privacy}, Srivastava et al.~\cite{srivastav2019human}, and Butler et al.~\cite{butler2015privacy}. Obfuscation-based approaches use off-the-shelf object detectors to identify privacy attributes and modify or remove the detected regions. Ren et al.~\cite{ren2018learning} synthesize fake images in place of detected faces in action detection, and Zhang et al.~\cite{zhang2021multi} use semantic segmentation followed by blurring for video privacy preservation. 
 
Adversarial training approaches, such as those proposed by Pittaluga et al. and Xiao et al. for privacy preservation in images~\cite{pittaluga2019learning,xiao2020adversarial}, and a novel framework for privacy-preserving action recognition introduced by ~\cite{wu2020privacy,wu2018towards}, utilize a minimax optimization strategy where the action classification cost is minimized, while the privacy classification cost is maximized.  
MaSS~\cite{chen2022mass} employs a framework similar to Wu et al.~\cite{wu2020privacy}, but with the adaptation of a compound loss to selectively preserve specific attributes rather than eliminate them. STPrivacy~\cite{li2023stprivacy} enhances this general framework by incorporating a transformer anonymizing block to mask entire video tubelets. SPAct~\cite{dave2022spact} obfuscates  all spatial semantic information while keeping utility action information via minimax optimization. TeD-SPAD~\cite{fioresi2023ted} builds on SPAct wherein they use NT-Xent~\cite{chen2020simple} contrastive loss in the budget branch to mitigate spatial privacy leakage. Recently, \cite{luo2024differentially} showed differentially private training on video based action recognition. While that work bears similarities, their results are limited to a lenient privacy budget of $\epsilon \in \{5,10\}$. In this work we specifically target the feasibility of strongly private low $\epsilon$ ranges.
 
\paragraph{Differential Privacy in Machine Learning.}
 
Differential Privacy (DP)~\cite{dwork2006calibrating} is a gold standard technique\cite{de2022unlocking} for formalizing the privacy guarantees of algorithms operating on sensitive datasets and has been integrated into a variety of machine learning tasks, including image classification~\cite{abadi2016deep, papernot2018scalable, tramer2020differentially, shamsabadi2021losing, zhu2020private}, activity recognition~\cite{luo2023differentially}, language modeling~\cite{yu2021differentially, shi2021selective, dinh2023context}, graph learning~\cite{sajadmanesh2021locally,sajadmanesh2023gap,sajadmanesh2023progap}, and speech recognition~\cite{shamsabadi2022differentially}. The most popular method for training deep neural networks with DP is Differentially Private Stochastic Gradient Descent (DP-SGD)~\cite{abadi2016deep}, which masks the contribution of any single data point to the model updates by clipping per-sample gradients and then adding Gaussian noise to the sum of the clipped gradients. However, this added noise often hampers optimization, leading to performance degradation compared to non-private training~\cite{dormann2021not, kurakin2022toward}. To address this, various methods have been proposed to improve the utility of DP-SGD, including adaptive clipping~\cite{andrew2021differentially}, adaptive noise scaling~\cite{fu2022adap}, and gradient compression~\cite{wang2021datalens}.  
Alternatively, some works have proposed relaxed notions of DP to protect only a subset of the data or provide different privacy levels per each data attribute~\cite{jorgensen2015conservative, kotsogiannis2020one,shi2021selective}. In
contrast to these prior works, our work differs substantially
in scope and technical approach. Our differentially private training scheme operates on masks selectively on the fused synthetic and real-world data,  wherein we decompose each video into two parts (utility and non-anonymized tokens) and selectively apply DP only to the non-anonymized tokens.

\section{Anonymization}
\label{sec:anon}

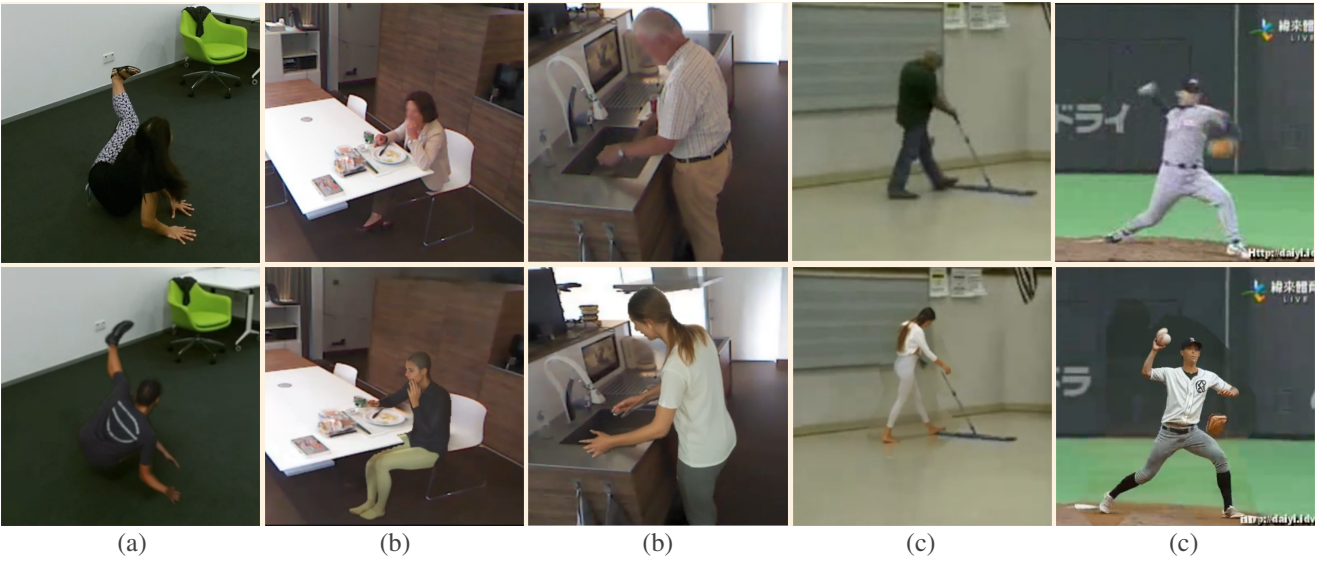
\begin{figure*}[!tb]
  \centering
    \begin{tikzpicture}
        \node[anchor=south west,inner sep=0pt] (image) at (0,0) {\includegraphics[width=\textwidth]{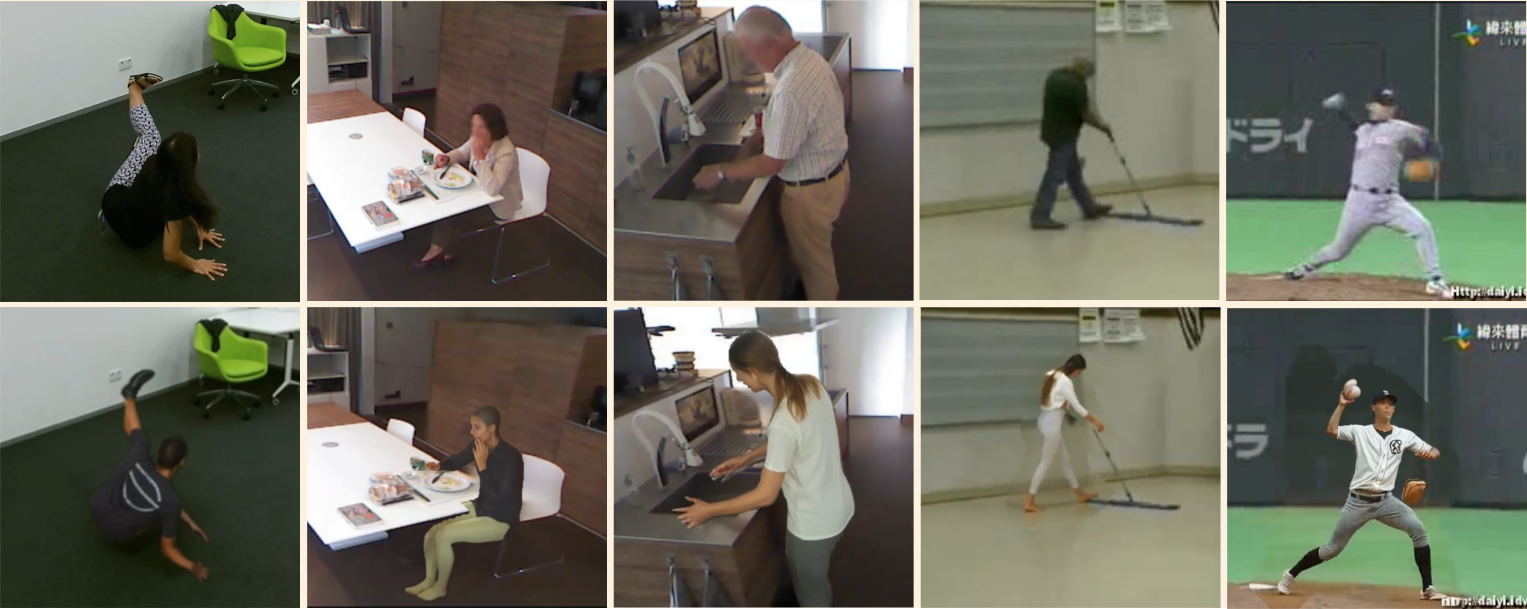}};
    \begin{scope}[x={(image.south east)}, y={(image.north west)}]
        \begin{scope}[anchor=center, text=gray!150]
            \node            at (.10, -.035) {(a)};
            \node            at (.30, -.035) {(b)};
            \node            at (.50, -.035) {(b)};
            \node            at (.70, -.035) {(c)};
            \node            at (.90, -.035) {(c)};
        \end{scope}
    \end{scope}
    \end{tikzpicture}
    \vspace{-2.2em}
  \caption{Anonymization examples from \emph{NTU RGB+D 60} (a), \emph{Toyota Smarthome} (b) and \emph{UCF-101} (c). Our anonymization method can be applied to difficult poses (a) and maintains important context information such as food on a table or a kitchen environment (b).}
  \label{fig:examples}
\end{figure*}

This section presents our video anonymization approach that effectively preserves individual privacy while maintaining video quality and contextual integrity. Our methodology integrates advanced pose estimation, 3D avatar rendering, and diffusion models to achieve high-fidelity, privacy-preserving video anonymization, as illustrated in Figure~\ref{fig:framework} and delineated in Algorithm~\ref{alg:SingleVideoAnonymization}.

The process initiates with the application of a pose estimation network $\PoseEstimation$~\cite{goel2023humans} on a video ${x}$ to extract temporally tracked 3D pose information, denoted as $\Pose$. Subsequently, our custom rendering function $\Rendering$ synthesizes diverse virtual humans using the SMPL-H model~\cite{SMPL:2015, MANO:SIGGRAPHASIA:2017}. The output of $\Rendering(\Pose)$ comprises rendered avatars $\RenderedPerson$, corresponding depth maps, and image masks $\RegionsAnonymized$ for the rendered subjects. Concurrently, we generate a person mask $\RegionsPerson$ utilizing the segmentation network $\PersonSegmentation$~\cite{li2022mask}. To ensure temporal consistency, we apply operation $B$ to $\RegionsPerson$; unflickering the masks accross frames while at the same time growing them slightly and applying a gaussian kernel for alpha blending on edges.

An inpainting mechanism $\VideoInpainting$~\cite{liCvpr22vInpainting} is applied on $x$ to remove any originally visible person, using the person mask $\RegionsPerson$, resulting in $\SampleWithPersonsRemoved$, followed by a merging of the resulting video and the generated synthetic avatars, yielding $\ddot{x}$. 

A temporally smooth tubelet video crop operation is then executed on $\ddot{x}$, $\RegionsAnonymized$, and $\RegionsPerson$, yielding one or multiple tubelets $\ddot{x}_j$, $\RegionsAnonymized_j$, and $\RegionsPerson_j$. For each tubelet crop $\ddot{x}_j$, we perform depth detection $E(\dot{x}_j)$, pose detection $P(\ddot{x}_j)$, and face keypoint detection $F(\ddot{x}_j)$, resulting in $C_d$, $C_p$, and $C_f$, respectively.

Our diffusion function $G$ integrates a stable diffusion network with AnimateDiff~\cite{guo2023animatediff} and three pre-trained controlnets. It processes each tubelet crop $\ddot{x}_j$, conditioning the model on depth ($C_d$), pose ($C_p$), and face keypoint ($C_f$) information. To accommodate unbound input video lengths within AnimateDiff's frame-limited approach, we employ an iterative frame sampling strategy. This approach constructs frame sets at varying strides, averaging latent representations during denoising, and applies to both input and conditioning frames as well as the diffusion process itself. The result is a refined, anonymized human figure $\hat{x}_j$, achieving high-quality video anonymization that balances realism, detail, and privacy preservation.

Finally, the anonymized humans $\{\hat{x}_j\}_{j=1}^M$ are reintegrated with the sanitized background $\SampleWithPersonsRemoved$, using the processed person masks $\{\RegionsPerson_j\}_{j=1}^M$, resulting in the fully anonymized video sample $\AnonymizedVideoSample$ as displayed at the bottom row of Figure~\ref{fig:examples}.

This process is repeated for each video in the dataset, resulting in an anonymized video dataset $\AnonymizedVideoDataset = \{(\AnonymizedVideoSample_1, \RegionsAnonymized_1, \LabelVector_1), \dots, (\AnonymizedVideoSample_N, \RegionsAnonymized_N, \LabelVector_N)\}$. Since $\RegionsAnonymized_i$ provides pixel-wise exact information of the presence of synthetic data, spatio-temporal regions which are masked as such can be considered completely anonymous and do not need additional privacy measures, while non-synthesized regions undergo masked differentially private training (\Cref{sec:dp}). 

There are limitations to this concept of anonymization which are addressed in Section~\ref{sec:limits} and can be found in more detail in the supplementary as well as additional implementation details and a large randomly selected set of anonymized videos for qualitative evaluation.

\begin{algorithm}[t] \DontPrintSemicolon \small 
\caption{Diffusion-refined synthetic humans}\label{alg:SingleVideoAnonymization} 
\KwInput{Raw Video $\RawVideoSample$; Pose estimation $\PoseEstimation$; Rendering $\Rendering$; Person segmentation $\PersonSegmentation$; Video inpainting $\VideoInpainting$} 
\KwResult{Anonymized Video $\AnonymizedVideoSample$} 

$\Pose \gets \PoseEstimation(\RawVideoSample)$  \Comment*{Estimate 4D poses} 
$\RenderedPerson, \RegionsAnonymized \gets \Rendering(\Pose)$ \Comment*{Generate avatars and render masks} 
$\RegionsPerson \gets \PersonSegmentation(\RawVideoSample)$  \Comment*{Generate person masks} 

$\RegionsPerson \gets B(\RegionsPerson)$ \Comment*{Unflicker and smooth masks} 

$\SampleWithPersonsRemoved \gets \VideoInpainting(\RawVideoSample, \RegionsPerson)$  \Comment*{Inpaint to remove persons} 

$\ddot{x} \gets \text{Merge}(\SampleWithPersonsRemoved, h,r)$  \Comment*{Overlay avatars} 

$\{\ddot{x}_{j}, \RegionsAnonymized_{j}, \RegionsPerson_{j}\}_{j=1}^M \gets C(\SampleWithPersonsRemoved, \RegionsAnonymized, \RegionsPerson)$ \Comment*{Smooth video crops} 

$\hat{x} \gets []$ \For{$j \gets 1$ \KwTo $M$}{ 
\tcp{Depth, pose, face keypoint detection}
$C_d,C_p,C_f \gets E(\ddot{x}_{j}),P(\ddot{x}_{j}),F(\ddot{x}_{j})$ \\
$\hat{x}_{j} \gets G(\ddot{x}_{j}, C_d, C_p, C_f)$ \Comment*{Video diffusion}  
$\hat{x} \gets \hat{x} \cup \{\hat{x}_{j}\}$ } 
$\AnonymizedVideoSample \gets K(\SampleWithPersonsRemoved, \{\hat{x}_j\}_{j=1}^M, \{\RegionsPerson_{j}\}_{j=1}^M)$ \Comment*{Reintegrate} 
\Return $\AnonymizedVideoSample, \RegionsAnonymized$ \end{algorithm}

\section{Masked Differentially Private Training}\label{sec:dp}

\newcommand{\Dataset}{\mathcal{D}}
Differential privacy (DP)~\cite{dwork2006calibrating} serves as a gold standard for ensuring the privacy of individuals' sensitive data.  
In simple terms, DP ensures that the output of an algorithm remains consistent regardless of whether a specific individual's data is included in the dataset. This means that an external observer who has access to the entire dataset except for one individual's information cannot determine with high probability whether that specific individual's data is part of the input. We provide a formal definition of DP below.
\begin{definition}[Differential Privacy~\cite{dwork2006calibrating}]\label{def:dp}
	Given $\epsilon > 0$ and $\delta > 0$, a randomized algorithm $\mathcal{A}$ satisfies $(\epsilon,\delta)$-differential privacy, if for all possible pairs of adjacent datasets $\Dataset$ and $\Dataset'$ differing by at most one record, denoted as $\Dataset\sim\Dataset'$, and for any possible set of outputs $S\subseteq Range(\mathcal{A})$, we have:
	\begin{equation*}
		\Pr[\mathcal{A}(\Dataset) \in S] \le e^\epsilon\Pr[\mathcal{A}(\Dataset') \in S] + \delta.
	\end{equation*}
\end{definition}
The parameter $\epsilon$ is referred to as the ``privacy budget'' or ``privacy cost,'' which balances the trade-off between privacy and utility. A smaller $\epsilon$ provides stronger privacy guarantees but might result in reduced utility. The parameter $\delta$ serves as a small failure probability, ensuring the robustness of the privacy mechanism. In the context of deep learning, DP techniques are employed during the model training process to protect the privacy of training data. The most popular algorithm, Differentially Private Stochastic Gradient Descent (DP-SGD)~\cite{abadi2016deep}, involves adding carefully calibrated noise to the clipped gradients computed during training, obscuring the contribution of individual data points.

\begin{algorithm}[t]
    \small
    \DontPrintSemicolon 
    \caption{MaskDP-SGD}\label{alg:ModelTraining}
    \KwInput{Dataset $ \RawVideoDataset = \{(\RawVideoSample_1, \LabelVector_1), \dots, (\RawVideoSample_\DatasetSize, \LabelVector_\DatasetSize)\}$; 
    Task model $ \TaskModel $ with parameters $\ModelParams$; Video function $\Tokenizer$; Loss function $\LossFunction$; Number of epochs $\NumEpochs$; Batch size $\BatchSize$; Gradient clipping threshold $\GradientClipThreshold$; noise standard deviation $\DPNoiseVariance$; Learning rate $\LearningRate$;
    }
    
    \KwResult{MaskDP Trained model $ \TaskModel $}
    
    
    Initialize $\ModelParams$ \;
    
    \For{$\TrainingIter \in [\NumEpochs \cdot \frac{\DatasetSize}{\Batch}]$}{
        Sample a batch $\Batch_\TrainingIter$ by selecting each $\Index\in[\DatasetSize]$ independently with probability $\nicefrac{\BatchSize}{\DatasetSize}$ \;
    
        \For{$\Index \in \Batch_\TrainingIter$}{
          \Comment{Tokenize video into private and public tokens.}
          $ \BackgroundTokens^{(t,i)}, \PersonTokens^{(t,i)} \gets \Tokenizer(\AnonymizedVideoSample_\Index, \RegionsAnonymized_\Index) $ \;
          \Comment{Compute gradients for both token sets}
          $ \GradientUtil{\TrainingIter, \Index} \gets \nabla_\ModelParams\LossFunction(\TaskModel(\PersonTokens^{(t,i)}), \LabelVector_\Index) $ \;
          $ \GradientPriv{\TrainingIter,\Index} \gets \nabla_\ModelParams\LossFunction(\TaskModel(\BackgroundTokens^{(t,i)}), \LabelVector_\Index) $ \;
          \Comment{Clip gradients of private tokens}
          $ \GradientPrivClipped{\TrainingIter,\Index} \gets \GradientPriv{\TrainingIter,\Index} \cdot \min(1, \frac{\GradientClipThreshold}{\|\GradientPriv{\TrainingIter,\Index}\|_2})$ \;
        }
        \Comment{Compute total gradient, add noise, and update parameters}
        $\GradientUtil{\TrainingIter} \gets \sum_{\Index=1}^{|\Batch_\TrainingIter|} \GradientUtil{\TrainingIter, \Index} ,\quad\quad \GradientPrivClipped{\TrainingIter} \gets \sum_{\Index=1}^{|\Batch_\TrainingIter|} \GradientPrivClipped{\TrainingIter,\Index} + \mathcal{N}(0, \DPNoiseVariance^2\vec{I}) $\;
        $ \GradientTotal{\TrainingIter} \gets \frac{1}{|\Batch_\TrainingIter|} \Big(\GradientUtil{\TrainingIter} + \GradientPrivClipped{\TrainingIter} \Big)$\;
        $ \ModelParams \gets \ModelParams - \LearningRate \GradientTotal{\TrainingIter} $ \;
    }
    
    \Return $ \TaskModel $ \;
    \end{algorithm}

While differential privacy protects information of dataset records, defining the composition of a record significantly influences both privacy and utility. Previous methods in the field of computer vision apply DP on the individual image or video level. For masked differential privacy we assume additional knowledge about the spatial or temporal region of sensitive information \emph{within} a dataset sample and limit the application of differential privacy to that region. This allows us to decompose visual data into sets of tokens $\mathcal{T}_\text{pr}$ and $\mathcal{T}_\text{pu}$ where only $\mathcal{T}_\text{pr}$ are considered private while $\mathcal{T}_\text{pu}$ are non-sensitive public tokens.

Hence, we introduce Masked Differential Privacy (MaskDP) to formally analyze privacy guarantees of an algorithm that offers exclusive protection of specific components of individual data points, rather than the entire record. To this end, we first define the notion of masked adjacency, which is the key concept in our definition of MaskDP:

\begin{definition}[Masked Adjacency]\label{def:maskedadj}
    Let $\Dataset=\{\Record^{(1)}, \dots, \allowbreak\Record^{(\DatasetSize)}\}$ be a dataset of size $\DatasetSize$, where each record $\Record^{(i)} = \{\Record^{(i)}_1,\dots,\Record^{(i)}_\NumTokens\}$ is partitioned into $\NumTokens$ tokens. Let $\MaskFunction: [N]\times[K]\rightarrow\{0,1\}$ be a mask function indicating which tokens are protected, where $\MaskFunction(i,k)=1$ if and only if $\Record^{(i)}_k$ is protected. Consider another dataset $\Dataset'=\{\Record'^{(1)},\dots,\Record'^{(\DatasetSize)}\}$, and for all $i\in\{1,\dots,\DatasetSize\}$, let $\SensitiveSet_i = \{ k\in[\NumTokens] : \MaskFunction(i,k)=1\}$. 
    Then $\Dataset$ is masked-adjacent to $\Dataset'$, denoted as $\Dataset\Adjacent\Dataset'$, if and only if $\exists i\in[\DatasetSize]$ such that $\forall j\in[\DatasetSize], j\ne i: \Record^{(j)}=\Record'^{(j)}$ and $\forall k\in[\DatasetSize]\setminus\SensitiveSet_i: \Record^{(i)}_k=\Record'^{(i)}_k$ and $\exists l\in\SensitiveSet_i$ such that $\Record^{(i)}_l\ne\Record'^{(i)}_l$.
\end{definition}

\setlength{\tabcolsep}{4pt}
\begin{table*}[htb!]
\caption{Accuracy for NTU RGB+D and mean Per Class Accuracy for Toyota Smarthome contrasting the model's performance with synthetic and anonymized data under Differential Privacy (DP) and Masked Differential Privacy (MaskDP) across various epsilon values.}
\label{tab:dp_ntu_tsh}
\centering
\begin{threeparttable}
\begin{tabular}{p{1.5em}|>{\raggedright\arraybackslash}p{4.3em}>{\centering\arraybackslash}p{4.2em}|rrrrrrr}
\toprule
& \multirow{2}{*}{Data} & \multirow{2}{*}{Privacy} & \multicolumn{5}{c}{\{Acc,mPCA\}@$\epsilon$} \\
&            &             & \cml{$0.1$}           & \cml{$0.25$}            & \cml{$0.5$}             & \cml{$0.75$}             & \cml{$1$}        & \cml{$5$}         & \cml{$\infty$} \\
\cmidrule{1-10}
\multirow{6}{*}{\rotatebox[origin=c]{90}{\centering NTU RGB+D 60}} 
& Real & \textsc{DP} & $1.9$~\ch{}             & $2.4$~\ch{}               & $15.9$~\ch{}              & $49.9$~\ch{}             & $55.2$~\ch{}            & $\textbf{70.8}$~\ch{} & $\textbf{83.8}$~\ch{} \\
& Render     & \textsc{DP} & $1.7$~\ch{-0.2}         & $2.3$~\ch{-0.1}           & $14.3$~\ch{-1.6}          & $30.7$~\ch{-19.2}         & $39.2$~\ch{-19}          & $52.9$~\ch{-17.9}     & 68.3~\ch{-15.5} \\
& Render     & MaskDP      & 4.2~\ch{2.3}            & $16.2$~\ch{13.8}          & 35.5~\ch{19.6}            & $48.3$~\ch{1.6}           & $51.0$~\ch{-4.2}         & $58.0$~\ch{-12.8}     & 68.3~\ch{-15.5} \\
& Rend.+Diff & \textsc{DP} & $2.1$~\ch{0.2}          & $2.8$~\ch{0.4}            & $34.5$~\ch{18.6}          & $51.2$~\ch{1.3}           & $56.3$~\ch{1.1}          & $65.0$~\ch{-5.8}      & $75.4$~\ch{-8.4} \\
& Rend.+Diff & MaskDP      & $\textbf{8.1}$~\ch{6.2} & $\textbf{19.9}$~\ch{17.5} & $\textbf{48.6}$~\ch{32.7} & $\textbf{54.0}$~\ch{4.1} & $\textbf{60.0}$~\ch{4.8} & $64.9$~\ch{-5.9}      & $75.4$~\ch{-8.4} \\
\cmidrule{2-10}
& \textit{Real$^\star$} & \textit{MaskDP$^\star$} & \textit{4.6}~\ch{} & \textit{26.6}~\ch{} & \textit{60.8}~\ch{} & \textit{61.0}~\ch{} & \textit{72.1}~\ch{} & \textit{77.3}~\ch{} & \textit{83.8}~\ch{} \\
\midrule
\multirow{6}{*}{\rotatebox[origin=c]{90}{\centering TSH}}
& Real & \textsc{DP} & $2.6$~\ch{}             & $3.8$~\ch{}               & $8.0$~\ch{}             & $14.0$~\ch{}              & $20.2$~\ch{}             & $29.9$~\ch{}             & $\textbf{52.7}$~\ch{} \\
& Render     & \textsc{DP} & $3.0$~\ch{0.4}          & $3.6$~\ch{-0.2}           & $4.4$~\ch{-3.6}         & $4.7$~\ch{-9.3}           & $7.3$~\ch{-12.9}         & $16.9$~\ch{-13.0}        & $26.7$~\ch{-26.0} \\
& Render     & MaskDP      & $3.2$~\ch{0.6}          & $9.1$~\ch{5.3}            & $11.1$~\ch{7.3}         & $14.4$~\ch{0.4}           & $16.9$~\ch{-3.3}         & $17.1$~\ch{-12.8}        & $26.7$~\ch{-26.0} \\
& Rend.+Diff   & \textsc{DP} & $4.0$~\ch{1.4} & $7.0$~\ch{3.5}            & $14.9$~\ch{11.1}        & $23.8$~\ch{9.8}           & $\textbf{28.0}$~\ch{7.8} & $\textbf{32.5}$~\ch{2.6} & $39.5$~\ch{-13.2} \\
& Rend.+Diff   & MaskDP      & $\textbf{12.4}$~\ch{9.8}               & $\textbf{15.4}$~\ch{11.6} & \textbf{22.3}~\ch{18.5} & $\textbf{24.9}$~\ch{10.9} & $25.6$~\ch{5.4}          & $27.8$~\ch{-2.1}                & $39.5$~\ch{-13.2} \\
\cmidrule{2-10}
& \textit{Real$^\star$} & \textit{MaskDP$^\star$} & \textit{3.9}~\ch{} & \textit{15.3}~\ch{} & \textit{18.9}~\ch{} & \textit{24.3}~\ch{} & \textit{27.3}~\ch{} & \textit{32.3}~\ch{} & \textit{52.7}~\ch{} \\
\bottomrule
\end{tabular}
\begin{tablenotes}
\item[$\star$] \textit{Real human visible, non-private, listed to show domain gap.}
\end{tablenotes}
\end{threeparttable}
\end{table*}

Intuitively, when two datasets $\Dataset$ and $\Dataset'$ are masked adjacent, it implies that they are almost identical except in one record, where a change occurs only in the tokens that are specifically marked as sensitive by the mask function $\MaskFunction$. 

Based on the above definition, we can now formally define MaskDP:
\begin{definition}[Masked Differential Privacy (MaskDP)]
    Given $\epsilon > 0$, $\delta > 0$, and mask function $\MaskFunction$, a randomized algorithm $\mathcal{A}$ satisfies $(\epsilon,\delta)$-MaskDP, if for all possible pairs of masked adjacent datasets $\RawVideoDataset\Adjacent\RawVideoDataset'$ under $\MaskFunction$, and for any possible set of outputs $S\subseteq Range(\mathcal{A})$, we have:
	\begin{equation*}
		\Pr[\mathcal{A}(\RawVideoDataset) \in S] \le e^\epsilon\Pr[\mathcal{A}(\RawVideoDataset') \in S] + \delta.
	\end{equation*}
\end{definition}

This definition formalizes the idea that privacy protection targets specific components within individual records, as opposed to protecting each record in its entirety. 


\paragraph{Remark.}Note that the only distinction between standard DP and MaskDP lies in the concept of dataset adjacency. As a result, fundamental properties of DP, including post-processing, composition, and privacy amplification theorems~\cite{dwork2006calibrating,abadi2016deep,wang2019subsampled}, also hold for MaskDP as they do not depend on any specific definition of adjacency. Moreover, the above definition readily extends to other variants of DP, such as Renyi DP~\cite{mironov2017renyi}, only by considering \Cref{def:maskedadj} as the notion of adjacency between datasets.

\paragraph{Learning with MaskDP}

\Cref{alg:ModelTraining} presents our proposed framework for implementing MaskDP-SGD, designed for training robust privacy-preserving neural networks on sensitive datasets. 

The algorithm computes the gradients of the model's loss function for both $\PersonTokens$ and $\BackgroundTokens$ token sets, separately. Only the gradients of background tokens $\BackgroundTokens$ are clipped to bound their contribution to the model update, followed by the addition of Gaussian noise.

The following theorem analyzes the privacy cost of our proposed algorithm. The proof is deferred to the supplementary.

\begin{theorem}\label{thm:privacy}
	Given a dataset $\Dataset$ of size $\DatasetSize$, batch-size $\BatchSize<\DatasetSize$, number of training epochs $\NumEpochs$, gradient clipping threshold $\GradientClipThreshold>0$, and Gaussian noise standard deviation $\DPNoiseVariance>0$, \Cref{alg:ModelTraining} satisfies $(\epsilon, \delta)$-MaskDP for any given $\delta\in(0,1)$, where:
      \begin{multline*}
        \epsilon \le \min_{\alpha>1} \nicefrac{\NumEpochs\DatasetSize}{\BatchSize(\alpha-1)}\log\left\lbrace \left(1-\SamplingProb\right)^{\alpha-1}\left( \alpha\SamplingProb - \SamplingProb + 1 \right) \right. \\
         \left. + \binom{\alpha}{2}\left(\SamplingProb\right)^2\left(1-\SamplingProb\right)^{\alpha-2}e^{\nicefrac{C^2}{\sigma^2}} \right. \\
         \left. +\sum_{l=3}^\alpha\binom{\alpha}{l}\left(1-\SamplingProb\right)^{\alpha-l}\left(\SamplingProb\right)^l e^{(l-1)(\nicefrac{C^2 l}{2\sigma^2})} \right\rbrace \\
          + \log\left(\nicefrac{\alpha - 1}{\alpha}\right) - \nicefrac{(\log \delta + \log \alpha)}{(\alpha - 1)},
      \end{multline*}
providing that the labels are not protected, i.e., assuming the label to be the first token of every record, then $\forall i\in [N], \MaskFunction(i,1)=0$.
\end{theorem}
\vspace{-1em}

\section{Experiments}\label{sec:setup}
\begin{table*}[htb!]
 	\caption{Accuracy on UCF-101 contrasting the model's performance with synthetic and anonymized data under Differential Privacy (DP) and Masked Differential Privacy (MaskDP) across various epsilon values.}
	\label{tab:dp_ucf}
	\centering
	\begin{threeparttable}
\begin{tabular}{p{1.5em}|>{\raggedright\arraybackslash}p{4.3em}>{\centering\arraybackslash}p{4.2em}|rrrrrr}
  \toprule
  & \multirow{2}{*}{Data} & \multirow{2}{*}{Privacy} & \multicolumn{6}{c}{Acc@$\epsilon$}                                     \\
  &                 &            & \cml{$0.25$}          & \cml{$0.5$}       & \cml{$0.75$}          & \cml{$1$}    & \cml{$5$}  & $\infty$~\ch{} \\
  \cmidrule{1-9}
  \multirow{6}{*}{\rotatebox[origin=c]{90}{\centering UCF-101}}
 & Real~\cite{luo2024differentially} & \textsc{DP} & \multicolumn{4}{c}{------}                                  & $75.1$~\ch{}     & ---~\ch{}    \\
 & Real       & \textsc{DP} & $1.4$~\ch{} & $2.7$~\ch{}      & $28.4$~\ch{}     & $77.1$~\ch{}      & $93.1$~\ch{}      & $97.6$~\ch{}      \\
 & Render     & \textsc{DP} & $1.6$~\ch{0.2} & $2.4$~\ch{-0.3}  & $6.8$~\ch{-21.6} & $57.4$~\ch{-19.7} & $89.0$~\ch{-4.1}  & $94.7$~\ch{-2.9}      \\
 & Render     & MaskDP      & $2.0$~\ch{0.6} & $19.0$~\ch{16.3} & $38.5$~\ch{10.1} & $58.4$~\ch{-18.7} & $85.3$~\ch{-7.3}  & $94.7$~\ch{-2.9}     \\
 & Rend.+Diff & \textsc{DP} & $1.0$~\ch{-0.4} & 2.0~\ch{-0.7}    & $10.4$~\ch{-18.0}  & 57.8~\ch{-19.3}   & 92.0~\ch{-1.1}    & 95.4~\ch{-2.2}    \\
 & Rend.+Diff & MaskDP      & $8.6$~\ch{7.2} & 36.2~\ch{33.5}   & $59.1$~\ch{30.7} & 72.9~\ch{-4.2}    & $91.0$~\ch{-2.1}  & 95.4~\ch{-2.2}   \\
  \bottomrule
\end{tabular}
	\end{threeparttable}
\vspace{-1em}
\hfill
\end{table*}
In this section, we evaluate MaskDP in comparison to standard differentially private training (DP-SGD) on synthetically anonymized datasets which only require region specific privacy. Since our work makes use of differential privacy with provable privacy guarantees on sensitive spatio-temporal regions, we follow similar work~\cite{abadi2016deep,luo2024differentially} and focus our experiments on the utility privacy tradeoff given a certain privacy budget $\epsilon$, rather than trying to approximate effective privacy with simulated attacks. We refer to the supplementary where we list additional experiments.

\subsection{Datasets}
Our experiments target multiple diverse and common human activity recognition datasets. NTU RGB+D~\cite{shahroudy2016ntu} is a balanced dataset containing 56K video samples labeled with 60 action categories and recorded in a very controlled setup. Toyota Smarthome~\cite{das2019toyota} comprises over 16K video clips containing 31 activities of daily living performed by multiple actors within an apartment which results in background and context information being descriptive of the performed activity. UCF-101\cite{soomro2012ucf101} is a comparatively small activity recognition datasets containing videos of actions performed in complex environments.
Utility on NTU RGB+D as well as UCF-101 is measured by classification accuracy, Toyota Smarthome is an unbalanced dataset and evaluation is measured by mean per-class accuracy (mPCA).

We apply our anonymization pipeline to these datasets as described in Section~\ref{sec:anon}. This process replaces real human subjects with synthetically-generated avatars while maintaining the original background and context. This allows us to create partially synthetic datasets where synthetically generated data is considered non-sensitive. To ablate the diffusion step in our pipeline we provide experiments for both, rendererd synthetic avatars as well as the additional refinement, referencing to this data as either \emph{Render} or \emph{Rend+Diff} in Tables~\ref{tab:dp_ntu_tsh} and ~\ref{tab:dp_ucf} in contrast to the unmodified data (\emph{Real}). Note, that in this setup it is the background information which is considered sensitive due to its insight about the living environment of a specific subject. This is in contrast to approaches which ignore sensitive information contained in background environments. 

\subsection{Implementation Details}
We utilize MViTv2-S\cite{li2021mvitv2} for action classification with Kinetics-400~\cite{kay2017kinetics} pre-trained weights, selectively fine-tuned using DP-SGD, similar to \cite{luo2024differentially}. This approach maintains higher utility than full fine-tuning, as DP-SGD may erode prior knowledge. MViTv2S, a Transformer model, uses LayerNorm instead of BatchNorm and meets the requirements of differential privacy. Experiments are conducted with batch size 128, employing (Masked) DP-SGD with a clipping threshold of 1 and a 0.01 learning rate over 150 epochs for NTU RGB+D and Toyota Smarthome and 300 epochs for UCF-101. The training includes a 20-epoch warmup followed by cosine decay. Non-private training uses standard SGD with identical parameters.

Data augmentation is used in non-private training but was excluded in our privacy-focused training since the noise added during training already acts as a regularizer and especially with low-$\epsilon$ settings training might not converge. We set $\delta$ for (Masked) DP-SGD at $10^{-6}$ for NTU RGB+D and $10^{-5}$ for Toyota Smarthome and UCF-101. 
We use Opacus to implement (Masked) DP-SGD and report results for multiple $\epsilon$ in all DP-related experiments.
For each $\epsilon$ value, we calibrate the noise multiplier to achieve the desired privacy level.
Further information on implementation details are listed in the supplementary.

\subsection{The Importance of Context}\label{sec:baselines}
\begin{table}[ht]
\centering
\caption{Performances on NTU RGB+D, Toyota Smarthome (TS) and UCF-101 without differentially private training.}
\label{tab:base}
\begin{tabular}{lcccc}
\toprule
\multirow{2}{*}{Description}  & \multirow{2}{*}{Privacy} & NTU & TS & UCF\\
             &         & \small{Acc} & \small{mPCA} & \small{Acc} \\
\midrule
Real World      & None   & $83.8$ & $52.7$ & $97.6$ \\
\midrule
Isolated Avatar    & Full   & $54.8$ & $8.8$  & $46.8$ \\
Inpainted Background & Person & $9.9$  & $10.6$ & $93.0$ \\
\midrule
Render on Background & Person & $68.3$ & $26.7$ & $94.7$ \\
RoB + Diffusion   & Person & $75.4$ & $39.5$ & $95.4$ \\
\bottomrule
\end{tabular}
\end{table}
\begin{figure}[!htb]
  \centering
  \includegraphics[width=\linewidth]{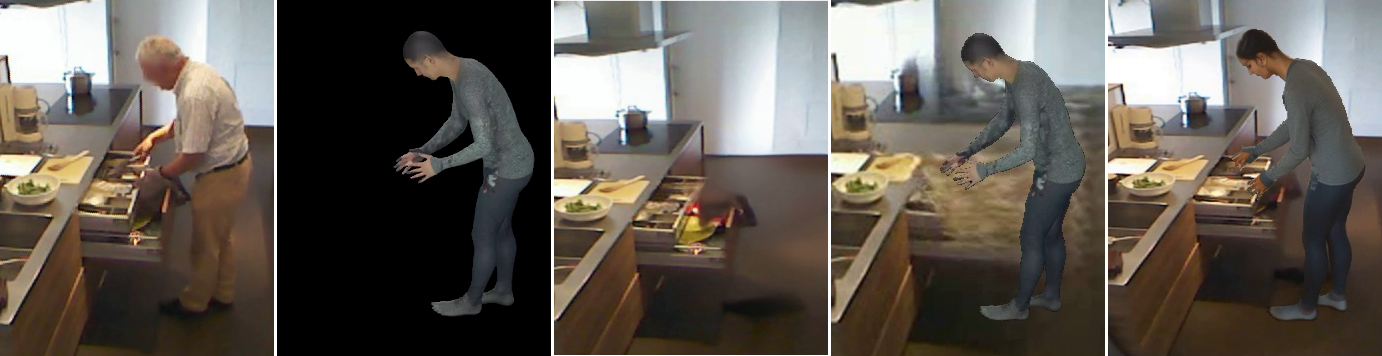}
  \caption{Examples for the data used in each row of Table~\ref{tab:base}.}
  \label{fig:abl_examples}
  \vspace{-1em}
\end{figure}

\begin{figure*}[!tb]
  \centering
  \includegraphics[width=\textwidth]{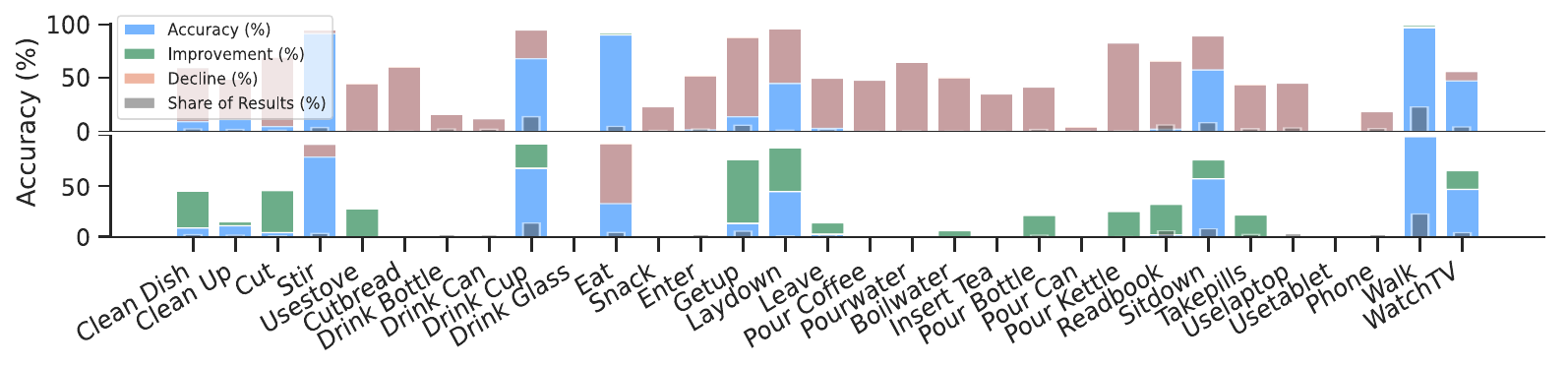}
  \vspace{-2em}
  \caption{
Class-wise accuracy loss on Toyota Smarthome compared to real world non-private training when applying DP (upper) and regain of accuracy upon applying MaskDP (lower) on anonymized data for $\epsilon = 5$. Class distribution marked in dark grey bars within.}
  \label{fig:tsh_classes}
\end{figure*}
To quantify our anonymization pipeline's impact and to assess contextual importance, we analyze performance across NTU RGB+D 60 (NTU), Toyota Smarthome (TS), and UCF-101 (UCF) datasets without the application of differential privacy (\Cref{tab:base}, examples in \Cref{fig:abl_examples}). Real-world baselines establish upper bounds, with TS exhibiting the highest complexity. Isolated avatars preserve privacy but significantly compromise utility, particularly in TS (8.8\% accuracy). NTU's controlled environment with uniform office background mitigates this effect (54.8\% accuracy), benefiting moderately from contextual information. Notably, inpainted backgrounds alone maintain high accuracy only in UCF (93.0\%), indicating its strong context dependence and suggesting heavy reliance on contextual cues.

Rendering avatars on real backgrounds substantially improves performance: NTU (68.3\% vs 54.8\%), TS (26.7\% vs 8.8\%), UCF (94.7\% vs 46.8\%). Incorporating our diffusion technique further enhances results, with TS showing the most significant gain (39.5\%, +12.8\%). These improvements underscore the critical role of real-world context, motivating our approach to combine synthetic avatars with real world backgrounds. Especially on NTU and UCF we reach full person anonymization while achieving a comparatively small synthetic-to-real utility domain gap. For setups which consider background information non-sensitive, our approach can directly be applied, for setups in which context itself is considered private we refer to our MaskDP experiments in Section~\ref{sec:exp_mdp}.

\subsection{MaskDP on Anonymized Activity Datasets}
\label{sec:exp_mdp}
In order to evaluate Masked Differential Privacy (MaskDP)  introduced in Section \ref{sec:dp},  we perform experiments with varying $\epsilon$ on combinations of fully real-world and person-anonymized data as well as normal differential privacy (DP) and MaskDP. We compare training on anonymized and non-anonymized datasets and report the non-private training results (indicated by $\epsilon=\infty$) as an upper bound. 
Table~\ref{tab:dp_ntu_tsh} displays our results on NTU RGB+D\cite{shahroudy2016ntu} as well as Toyota Smarthome~\cite{das2019toyota}.
Two competing effects are visible, a synthetic-to-real domain gap as analyzed in Section~\ref{sec:baselines} reducing performance by switching from purely real world to anonymized data as well as a loss of performance with progressively tighter privacy bound~$\epsilon$. 
Standard DP provides full differential privacy on a complete video clip, but suffers from prohibitive utility loss for $\epsilon \leq 1$ which is considered strongly private. For input data which is anonymized, MaskDP can be used, limiting DP application to sensitive image regions and significantly improving utility on both datasets.
To further analyze the effect if the domain gap on MaskDP results, we also applied MaskDP on non-anonymized data, utilizing the person masks from our anonymization pipeline and find that performance significantly increases in all settings, e.g. from 48.6~\% to 60.8~\% on NTU RGB+D for $\epsilon=0.5$, indicating that MaskDP profits from increasingly realistic data synthetisation methods. 

In Table~\ref{tab:dp_ucf} we list corresponding results for UCF-101. For $\epsilon$ $5$, this also allows us to compare our implementation to the experiments described in~\cite{luo2024differentially}. In our work we leverage MviTv2\cite{li2021mvitv2} while \cite{luo2024differentially} makes use MViT~\cite{fan2021multiscale}, as a result we are able to achieve significantly better results on the same baseline settings. \cite{luo2024differentially} provide additional results for their complementary method of multi-clip training, for comparability we refer to their single clip results. MaskDP on UCF-101 generally displays similar patterns to the experiments on NTU RGB+D and Toyota Smarthome, but limited to improvements within $\epsilon \leq 0.75$.

\section{Differential Privacy on Imbalanced Datasets}
\label{sec:dat_imb}
Differentially private training limits the impact of individual data samples on the final model, which can significantly impair performance on minority classes with few training samples. This is particularly evident in the imbalanced Toyota Smarthome dataset, as illustrated in \Cref{fig:tsh_classes} for $\epsilon = 5$.

Each bar contains a smaller dark gray bar representing the class's share of samples in the dataset. The upper graph shows DP performance in comparison to standard SGD training, while the lower graph shows MaskDP performance in comparison to DP training, thereby first displaying the detrimental effects of DP on minority classes and then presenting how MaskDP alleviates this effect.

DP training sacrifices performance on minority classes, maintaining good results on few majority classes like \emph{Stir}, \emph{Eat} and \emph{Walk}.
While MaskDP does not match the performance of SGD training, it demonstrates a notable ability to recover utility across multiple classes. This improvement is particularly visible for minority classes such as \emph{Usestove}, \emph{Pour Bottle}, and \emph{Takepills} which are not recognized at all with DP training.

\section{Limitations}
\label{sec:limits}
Anonymization with avatars induces a synthetic-to-real domain gap, limiting the use of MaskDP to strongly private setups. Further improvement of synthetic data generation methods can shift this turning point into higher $\epsilon$ regions. 

MaskDP provides differential privacy guarantees for spatio-temporal regions it covers. Non-masked regions lack such guarantees and might leak information if not specifically accounting for such risks. 

While differentially private training suffers in training from unbalanced datasets like Toyota Smarthome, we find that MaskDP can alleviate this effect to some degree. A further analysis and discussion of these limitations can be found in the supplementary.

\section{Conclusion}\label{sec:conclusion}
Our work addresses the crucial emerging problem of privacy-preserving action recognition. We introduce a novel approach which achieves high-quality anonymization of human activity datasets at scale without manual intervention and integrate it seamlessly with differentially private training, prioritizing privacy preservation while maintaining utility. Our key contributions include a framework for dataset anonymization as well as a new differentially private mechanism named MaskDP, operating selectively on fused synthetic and real-world data. Empirical evaluations show that our approach exhibits superior utility vs. privacy trade-offs in strongly private setups.

{
    \small
    \bibliographystyle{ieeenat_fullname}
    \bibliography{bibfiles/egbib,bibfiles/action_privacy,bibfiles/dp_in_ml,bibfiles/synthetic_action_datasets}
}

\newpage
\section{Code Release Statement}
The code for our dataset anonymization pipeline and Masked Differential Privacy implementation will be made publicly available with detailed documentation. During our work, we encountered challenges due to the lack of publicly accessible implementations in this field, which made it more difficult to build on existing methods. By releasing our code, we aim to address this gap and facilitate reproducibility and further research.

\section{Anonymized Dataset Samples}
With this supplementary material we provide 100 paired samples of real-world and anonymized videos, once for the NTU RGB+D and once for the Toyota Smarthome dataset. These samples facilitate direct qualitative evaluation of our anonymization pipeline by contrasting both the original and anonymized versions. Selection was performed randomly to ensure unbiased representation. Due to the class imbalance in the Toyota Smarthome dataset, this results in some actions appearing more frequently than others.

\section{Focus on Privacy-Preserving Models}
This work focuses on protecting private information in machine learning models trained on sensitive video data. Our primary goal is to ensure that the released model does not inadvertently leak sensitive information, aligning with differential privacy principles. This distinguishes our approach from those centered on dataset anonymization for public distribution~\cite{li2023stprivacy,dave2022spact}, as our focus is on safeguarding against attacks on a trained model.

To achieve this, we utilize an anonymization pipeline that replaces real human figures with synthetic avatars while retaining the original video context. This step ensures that sensitive features are concealed during training, preventing the model from internalizing private information. Combined with differential privacy mechanisms, such as MaskDP, this approach ensures that the exposure of privacy-sensitive details is minimized.

An inherent byproduct of this pipeline is that it generates avatar-anonymized datasets. This is a complementary outcome aligned with specific use cases where background information does not need protection.

\section{More Detailed Limitations}
\label{sup:limits}
Differentially private training assumes a utility-privacy trade-off controlled by the privacy budget $\epsilon$. Commonly, this budget is categorized into settings of strong formal privacy guarantees with $0 < \epsilon \leq 1$, reasonable formal privacy guarantees with $1 < \epsilon \leq 10$ and weak to no privacy guarantees with $10 < \epsilon$ \cite{ponomareva2023dp}. We make use of synthetic data as a replacement for differential privacy in selected regions of the input data, thereby foregoing the application of high noise on gradient updates influenced by these areas. While this is beneficial in strong privacy settings, this induces a synthetic-to-real domain gap which impacts utility and suggests the usage of vanilla differential privacy in higher-$\epsilon$ setups or the usage of better synthetization methods. 

Differential privacy induces utility penalties for unbalanced datasets such as Toyota Smarthome. MaskDP alleviates this effect but a significant utility gap remains. In \Cref{sec:dat_imb} we evaluate these limitations by looking at class-wise accuracies for this dataset. 

In \Cref{sec:dp} we define the concept of \emph{Masked Adjacency} in order to define our notion of differential privacy in MaskDP. Understanding the limitation provided by masked adjacency in contrast to whole sample based adjacency is important. In our setting we are only differentially private in terms of the sensitive private tokens, while we assume that non-sensitive regions do not contain private information and thereby do not require privacy measures. In our anonymization experiments we can ensure this since we have knowledge of pixelwise replacement with synthetic data. Note, that this certainty can not be guaranteed in many other setups. Consequently we do not consider our method end-to-end differentially private on \emph{whole} data samples. We still do consider it fully private since all remaining areas are perfectly anonymized in our experiments.

\section{Network Architecture}
For our experiments we make use of the MViTv2-S architecture, pretrained with Kinects-400. Although being similar, this setup is different to the one used in~\cite{luo2024differentially} which make use of MViT-B (v1). We follow them in training the normalization layers as well as the linear projection head, though we find that it is enough to train the normalization layers of the last two blocks, resulting in 50k trainable parameters vs the 110k trainable parameters referred to in~\cite{luo2024differentially}. This significantly increases training speed, since backpropagation propagates through the whole network until reaching the lowest trainable layer.

\section{Renyi Differential Privacy}\label{sec:rdp} 

In this work, we employ a variant of DP, called Renyi differential privacy (RDP)~\cite{mironov2017renyi} to analyze the privacy cost of our proposed method. RDP is a relaxation of DP, which provides a tighter privacy bound when composing multiple differentially private mechanisms. The following formally defines RDP:
\begin{definition}[Renyi Differential Privacy~\cite{mironov2017renyi}]\label{def:rdp}
	Given $\alpha > 1$ and $\epsilon > 0$, a randomized algorithm $\mathcal{A}$ satisfies $(\alpha,\epsilon)$-RDP if for every adjacent datasets $\Dataset$ and $\Dataset^\prime$, we have:
  \begin{equation}
    D_\alpha\left( \mathcal{A}(\Dataset) \Vert \mathcal{A}(\Dataset^\prime) \right) \le \epsilon,
  \end{equation}
	where $D_\alpha(P\Vert Q)$ is the {Rényi divergence} of order $\alpha$ between probability distributions $P$ and $Q$ defined as:
	\begin{equation*}
		D_\alpha(P\Vert Q) = \frac{1}{\alpha-1}\log\mathbb{E}_{x\sim Q}\left[\frac{P(x)}{Q(x)}\right]^\alpha.
	\end{equation*}
\end{definition}

Proposition~3 of~\cite{mironov2017renyi} analyzes the privacy cost of composing two RDP algorithms as the following:
\begin{proposition}
  [Composition of RDP mechanisms~\cite{mironov2017renyi}]
  \label{prop:rdpcompose}
    Let $f(\Dataset)$ be $(\alpha,\epsilon_1)$-RDP and $g(\Dataset)$ be $(\alpha,\epsilon_2)$-RDP, then the mechanism defined as $h(\Dataset) = \{f(\Dataset),g(\Dataset)\}$ satisfies $(\alpha,\epsilon_1+\epsilon_2)$-RDP.
  \end{proposition}

The following Proposition demonstrates how to convert RDP back to standard $(\epsilon,\delta)$-DP:
\begin{proposition}
[From RDP to $(\epsilon,\delta)$-DP~\cite{mironov2017renyi}]
\label{prop:rdptodp}
	If $\mathcal{A}$ is an $(\alpha,\epsilon)$-RDP algorithm, then it also satisfies $(\epsilon+\nicefrac{\log (1/\delta)}{\alpha-1},\delta)$-DP for any $\delta\in(0,1)$.
\end{proposition}

A better conversion law is provided in Theorem 21 of Balle et al.~\cite{balle2020hypothesis}:
\begin{theorem}
[Better RDP to $(\epsilon,\delta)$-DP conversion~\cite{balle2020hypothesis}]
\label{thm:rdptodp2}
	If $\mathcal{A}$ is $(\alpha,\epsilon)$-RDP, then it is $\left( \epsilon + \log\left(\nicefrac{\alpha - 1}{\alpha}\right) - \nicefrac{(\log \delta + \log \alpha)}{(\alpha - 1)}, \delta\right)$-DP for any $\delta\in(0,1)$.
\end{theorem}

As mentioned in \Cref{sec:dp}, we can easily define Masked RDP (MRDP) only by considering masked adjacency (\Cref{def:maskedadj}) as the notion of dataset adjacency in \Cref{def:rdp}. Consequently, \Cref{prop:rdpcompose,prop:rdptodp} also hold for Masked RDP as they are independent of any specific definition of adjacency.

\section{Deferred Theoretical Proofs}\label{sec:proof}

\subsection{Proof of \autoref{thm:privacy}}

To prove \Cref{thm:privacy}, we first need to establish the following lemma:

\begin{lemma}\label{lem:privacy}
    Given a dataset $\Dataset=\{\Record^{(1)},\dots,\Record^{(N)}\}$, where each record $\Record^{(i)}=\{\Record^{(i)}_1,\dots,\Record^{(i)}_\NumTokens\}$ is partitioned into $\NumTokens$ tokens, and a mask function $\MaskFunction:[N]\times[K]\rightarrow\{0,1\}$, let
    \[\ProtectedDataset=\left\{ \{ \Record^{(i)}_k \mid \forall k\in[\NumTokens]\text{ s.t. }\MaskFunction(i,k) = 1 \} \mid i\in[\DatasetSize] \right\}\]
    be a dataset consisting of only protected tokens. Correspondingly, let
    \[\UnprotectedDataset=\left\{ \{ \Record^{(i)}_k \mid \forall k\in[\NumTokens]\text{ s.t. }\MaskFunction(i,k) = 0 \} \mid i\in[\DatasetSize] \right\}\]
    be the set of unprotected tokens. Assume that $f_1(.)$ is an $(\alpha,\epsilon)$-RDP algorithm and $f_2(.)$ is an arbitrary function. Then the mechanism $g(.)$ defined as 
    \begin{equation}
        g(\Dataset)=\{f_1(\ProtectedDataset),f_2(\UnprotectedDataset)\}
    \end{equation}
    satisfies $(\alpha,\epsilon)$-MRDP under the same mask function $\MaskFunction$.
\end{lemma}

\begin{proof}
    Let $\Dataset$ and $\Dataset'$ be two masked-adjacent datasets the mask function $\MaskFunction$. The goal is to show 
    \begin{align*}
        D_\alpha\left( g(\Dataset) \Vert g(\Dataset^\prime) \right) \le \epsilon.
    \end{align*}
    Let $P_\Dataset$ and $P_{\Dataset'}$ denote the output distribution of $g(\Dataset)$ and $g(\Dataset')$, respectively. Then, for each $x\in Range(g)$, we have:
    \begin{align}
        P_\Dataset(x) &= P_{\Dataset_p}(x_1)P_{\Dataset_u}(x_2) \label{eq:privacy1} \\
        P_{\Dataset'}(x) &= P_{\Dataset'_p}(x_1)P_{\Dataset'_u}(x_2), \label{eq:privacy2}
    \end{align}
    where $P_{\Dataset_p}$, $P_{\Dataset_u}$, $P_{\Dataset'_p}$, and $P_{\Dataset'_u}$ are the output distributions of $f_1(\Dataset_p)$, $f_2(\Dataset_u)$, $f_1(\Dataset'_p)$, and $f_2(\Dataset'_u)$, respectively, and 
    $x=(x_1,x_2)$, $x_1\in Range(f_1)$, and $x_2\in Range(f_2)$. Therefore, we have:
    \begin{align}
        D_\alpha&\left( g(\Dataset) \Vert g(\Dataset^\prime) \right) = \frac{1}{\alpha-1}\log\mathbb{E}_{x\sim P_{\Dataset'}}\left[\frac{P_\Dataset(x)}{P_{\Dataset'}(x)}\right]^\alpha \nonumber \\
        &= \frac{1}{\alpha-1}\log\mathbb{E}_{(x_1,x_2)\sim P_{\Dataset'}}\left[\frac{P_{\Dataset_p}(x_1)P_{\Dataset_u}(x_2)}{P_{\Dataset'_p}(x_1)P_{\Dataset'_u}(x_2)}\right]^\alpha \nonumber \\
        &= \frac{1}{\alpha-1}\log\mathbb{E}_{x_1\sim P_{\Dataset'_p}}\left[\frac{P_{\Dataset_p}(x_1)}{P_{\Dataset'_p}(x_1)}\right]^\alpha \le \epsilon. \nonumber
    \end{align}
    The second equality follows from \eqref{eq:privacy1} and \eqref{eq:privacy2}. 
    The third equality follows from the fact that $\UnprotectedDataset = \UnprotectedDataset'$, which itself follows from the definition of masked adjacency (\Cref{def:maskedadj}). 
    Finally, the last inequality follows from the fact that $f_1(.)$ is an $(\alpha,\epsilon)$-RDP algorithm.
\end{proof}

Now, we are ready to prove \Cref{thm:privacy}:

\begin{proof}
    Each iteration of \Cref{alg:ModelTraining} applies the Gaussian mechanism on the set of non-anonymized tokens, denoted as $\Dataset^{(t)}_p = \{\BackgroundTokens^{(t,i)}: \forall i\in\Batch_t\}$, which according to Corollary~3 of~\cite{mironov2017renyi}, is $(\alpha,\nicefrac{\alpha C^2}{2\sigma^2})$-RDP. It also uses the anonymized tokens for the parameter update. According to \Cref{lem:privacy}, these two steps together satisfy $(\alpha,\nicefrac{\alpha C^2}{2\sigma^2})$-MRDP w.r.t. the batch $\Batch$. As each batch is randomly subsampled, according to Theorem~5~of~\cite{zhu2019poission}, each iteration of \Cref{alg:ModelTraining} satisfies $(\alpha,\epsilon_t)$-MRDP w.r.t. the entire dataset, where $\epsilon_t$ is given by:
    \begin{multline}\label{eq:epsilon_t}
        \epsilon_t  \le \frac{1}{\alpha-1}\log\left\lbrace \left(1-\SamplingProb\right)^{\alpha-1}\left( \alpha\SamplingProb - \SamplingProb + 1 \right) \right. \\
        \left. + \binom{\alpha}{2}\left(\SamplingProb\right)^2\left(1-\SamplingProb\right)^{\alpha-2}e^{\frac{C^2}{\sigma^2}} \right. \\
        \left. +\sum_{l=3}^\alpha\binom{\alpha}{l}\left(1-\SamplingProb\right)^{\alpha-l}\left(\SamplingProb\right)^l e^{(l-1)(\frac{C^2 l}{2\sigma^2})} \right\rbrace.
    \end{multline}
    As the algorithm runs for $\nicefrac{\NumEpochs\DatasetSize}{\BatchSize}$ iterations, based on the composition theorem of RDP (Proposition~1 of~\cite{mironov2017renyi}), \Cref{alg:ModelTraining} satisfies $(\alpha,\nicefrac{\NumEpochs\DatasetSize\epsilon_t}{\BatchSize})$-MRDP.
    The proof is completed by replacing $\epsilon_t$ with \eqref{eq:epsilon_t} and applying \Cref{thm:rdptodp2}.
\end{proof}

\end{document}